%% file: main.tex
\documentclass[3p,12pt]{elsarticle}

\usepackage{amssymb}
\usepackage{amsmath,amsfonts}
\usepackage{url}
\usepackage{graphicx}
\usepackage{framed,multirow}
\usepackage{lineno}
\usepackage{color}
\usepackage{xcolor}
\usepackage{colortbl}
\usepackage{bbm}
\usepackage{subfig}

\journal{XXX}

\begin{document}
\begin{frontmatter}

\title{EsurvFusion: An evidential multimodal survival fusion model based on Gaussian random fuzzy numbers}

\author[inst1]{Ling Huang \corref{cor1}}
\affiliation[inst1]{organization={Saw Swee Hock School of Public Health, National University of Singapore, Singapore},
            }

\author[inst1]{Yucheng Xing \corref{cor1}}
\cortext[cor1]{These authors contributed equally to this work.}
\author[inst1]{Qika Lin}
\author[inst2]{Su Ruan}
\affiliation[inst2]{organization={Université de Rouen Normandie, Quantif, LITIS},
            city={Rouen},
            country={France}}
\author[inst1,inst3]{Mengling Feng}
\affiliation[inst3]{organization={Institute of Data Science, National University of Singapore, Singapore},          
            }

\begin{abstract}

Multimodal survival analysis aims to combine heterogeneous data sources (e.g., clinical, imaging, text, genomics) to improve the prediction quality of survival outcomes. However, this task is particularly challenging due to high heterogeneity and noise across data sources, which vary in structure, distribution, and context. Additionally, the ground truth is often censored (uncertain) due to incomplete follow-up data. In this paper, we propose a novel evidential multimodal survival fusion model, EsurvFusion, designed to combine multimodal data at the decision level through an evidence-based decision fusion layer that jointly addresses both data and model uncertainty while incorporating modality-level reliability. Specifically, EsurvFusion first models unimodal data with newly introduced Gaussian random fuzzy numbers, producing unimodal survival predictions along with corresponding aleatoric and epistemic uncertainties. It then estimates modality-level reliability through a reliability discounting layer to correct the misleading impact of noisy data modalities. Finally, a multimodal evidence-based fusion layer is introduced to combine the discounted predictions to form a unified, interpretable multimodal survival analysis model, revealing each modality’s influence based on the learned reliability coefficients. This is the first work that studies multimodal survival analysis with both uncertainty and reliability.
Extensive experiments on four multimodal survival datasets demonstrate the effectiveness of our model in handling high heterogeneity data, establishing new state-of-the-art on several benchmarks.  
\end{abstract}


\end{frontmatter}  
\input{sec/1_intro}
\input{sec/2_related}

\input{sec/3_method}

\input{sec/4_exp}
\input{sec/5_conclu}
{
    \small
    \bibliographystyle{ieeenat_fullname}
    \bibliography{main}
}

\end{document}

%% file: sec/1_intro.tex
\section{Introduction}
\label{sec:intro}

Existing studies have widely proved that combining information from different modalities can provide complementary information and be helpful for accurate decisions in high-risk applications, e.g., medical decision \cite{muhammad2021comprehensive, huang2025deep, jaume2024modeling, pmlr-v235-song24b}, autonomous driving \cite{caesar2020nuscenes, prakash2021multi}.

State-of-the-art multimodal fusion research mainly focuses on discrete prediction tasks such as classification. That is mainly because of the convenience in probability operations where the prediction uncertainty can be easily managed through Bayesian reference or ensembling. 
The continuous multimodal fusion model lacks study due to its sensitivity to data alignment, prediction errors, and uncertainty. For example, for the classification task, only the correct class among several options needs to be identified, making it less affected by small variations in probability output. Therefore, existing continuous multimodal fusion research mainly focuses on the learning of a unified representation space with technologies such as attention mechanisms \cite{chen2021multimodal, jaume2024modeling, pmlr-v235-song24b}. However, an effective representation space can only be obtained under the training of enough high-quality data. Moreover, those approaches often lack interpretability at the decision level as the contribution of each modality in relation to the predicted real line output is difficult to clearly understand. For classification tasks, interpretability can often be managed by observing which modality had the highest influence on predicting a class, making it generally easier to explain than continuous prediction tasks. Additionally, the uncertainty and reliability regarding multimodal data are seldom studied, further limiting their performance.

Survival analysis, also known as time-to-event analysis, focuses on predicting the time it takes for an event of interest to occur, e.g., time to death or disease recurrence. It is a special kind of regression model in which the data often includes a large amount of censored (uncertain) survival times due to early-ended studies or lack of follow-up, resulting in only partially certain ground truth, making the modeling of the survival function more challenging than typical regression tasks. A recently proposed survival analysis model ENNreg \cite{huang2024evidential, huang2024evidentialtimetoevent} handles both uncensored and censored data by using efficient evidence modeling based on Epstemic random fuzzy set \cite{denoeux2021belief}. This approach estimates the event time and corresponding uncertainty based on newly introduced Gaussian random fuzzy numbers (GRFNs) \cite{denoeux2023quantifying}. The GRFNs can be regarded as generalized Gaussian random variables with fuzzy mean, which stands out as a promising new function to quantify aleatory and epistemic uncertainty for real line prediction tasks. However, this approach focuses exclusively on unified data sources, without accounting for the heterogeneity and the reliability of multimodal survival data.  

Inspired by ENNreg with its nice attribute in handling data censoring and uncertainty quantification, in this paper, we focus on multimodal survival tasks and propose a new effective multimodal survival fusion model (EsurvFusion) that can address data censoring and uncertainty, as well as considering modality level uncertainty and reliability for final prediction fusion.
Firstly, we model unimodal data with the newly introduced GRFNs to output unimodal survival prediction along with the aleatoric and epistemic uncertainty. Secondly, we address reliability differences among multimodal data by introducing a reliability discounting layer to correct the misleading impact of noisy data modality. Finally, we propose a multimodal evidence fusion layer that combines the discounted prediction to form a unified multimodal survival analysis model. 
Our contributions can be summarized into four main points:
\begin{itemize}
\item We propose a novel evidential multimodal survival analysis model, EsurvFusion, which combines multimodal data at the final decision level.
\item We incorporate (i) unimodal survival prediction with aleatory and epistemic uncertainty using the newly introduced GRFN, (ii) reliability-adjusted evidence correction via possibility discounting, and (iii) an evidence fusion layer that combines predictions from different modalities.

\item we propose a hybrid loss training strategy to reduce the risk of misleading decisions when learning from highly heterogeneous multimodal data.

\item We demonstrate that (i) survival predictions are interpretable, with quantified possible survival times and corresponding aleatory and epistemic uncertainty, and (ii) the learned reliability coefficients offer insights into the contribution of each input modality in the continuous prediction process, enhancing transparency and interpretability
\end{itemize}
Through complementary experiments on one image-clinical and three clinical-genomic cancer survival datasets, we show that the proposed EsurvFusion improves survival prediction reliability and quality and is benchmarked against both single and multimodal fusion methods.

%% file: sec/2_related.tex
\section{Related Work}
\label{sec:formatting}


\subsection{Multimodal fusion}
Multimodal information fusion strategies can be implemented at the early, middle, and late stages \cite{hermessi2021multimodal, weng2024semi}. \emph{Early fusion} concatenates multimodal data as a single input. However, the fused outcomes are not always promising due to domain shifts \cite{qi2018unified}, information distribution discrepancies \cite{xiao2019multi}, information entropy differences \cite{xu2017information}, etc. \emph{Middle fusion} is typically performed within a neural network by learning a shared representation or a joint embedding layer for multimodal inputs \cite{liang2021attention}. Though middle fusion can mitigate the drawbacks of early fusion with richer multimodal representations, it is difficult to study the contribution of each single modality to the final decision.
\emph{Late fusion} aggregates modality-level decisions from independent decision-makers. Late fusion can be further grouped into hard and soft fusion methods \cite{hall2008framework, huang2025deep}. Hard fusion methods aggregate logical information membership values, such as model ensembling with average votings \cite{khan2024hybrid}, making it compatible with probabilistic models like deep neural networks. However, averaging predictions can introduce bias, giving some modalities more weight than warranted. Soft fusion, which adjusts contributions based on each modality’s confidence, e.g., fuzzy voting \cite{hall2008framework}, is more consistent with real-world conditions.

\subsection{Uncertainty quantification}
As pointed out in \cite{rogova2004reliability, dubois2016basic,pichon19}, the success of information fusion depends on the quality of input information, the accuracy of prior knowledge, and the effectiveness of the uncertainty model used. Given the first two terms intricately depend on the data collection stage, researchers are focusing on developing effective fusion methods with a primary emphasis on uncertainty quantification \cite{abdar2021review, huang2023review, gao2024embracing}.

Early methods of uncertainty quantification essentially relied on probabilistic models, often integrated with Bayesian inference or sampling techniques to estimate uncertainty across various parameters or variables \cite{hinton1993keeping}. The advent of deep neural networks has sparked renewed interest in uncertainty estimation \cite{abdar2021review}, leading to the development of methods such as Monte Carlo simulation \cite{gal2016dropout} and ensembling \cite{lakshminarayanan2017simple}. Instead of making strong assumptions on actual data distribution, nonprobabilistic methods such as possibility theory \citep{zadeh1978fuzzy} and Dempster-Shafer (DS) theory of evidence \cite{dempster1967upper, shafer1976mathematical} provide alternative mathematical frameworks to quantify uncertainty than the above probabilistic models. Possibility theory uses a possibility distribution to describe how plausible various outcomes are. DS theory provides a powerful framework for imperfect (noise, uncertainty, partial reliable) information modeling, reasoning, and fusion through Dempster’s combination rule \cite{denoeux20b, mercier2008refined, pichon19}.
However, due to the lack of practical models that can handle continuous variables while satisfying Dempster’s combination rule, most applications of the DS theory are based on discrete models \cite{xu2016evidential, zou2022tbrats, huang2023application}. As an extension of both DS theory and possibility theory in real lines, the newly introduced epistemic random fuzzy set (ERFS) theory \cite{denoeux2021belief} enables the study of continuous models for real line prediction tasks.

\begin{figure*}[!ht]
  \centering
   \includegraphics[width=\linewidth]{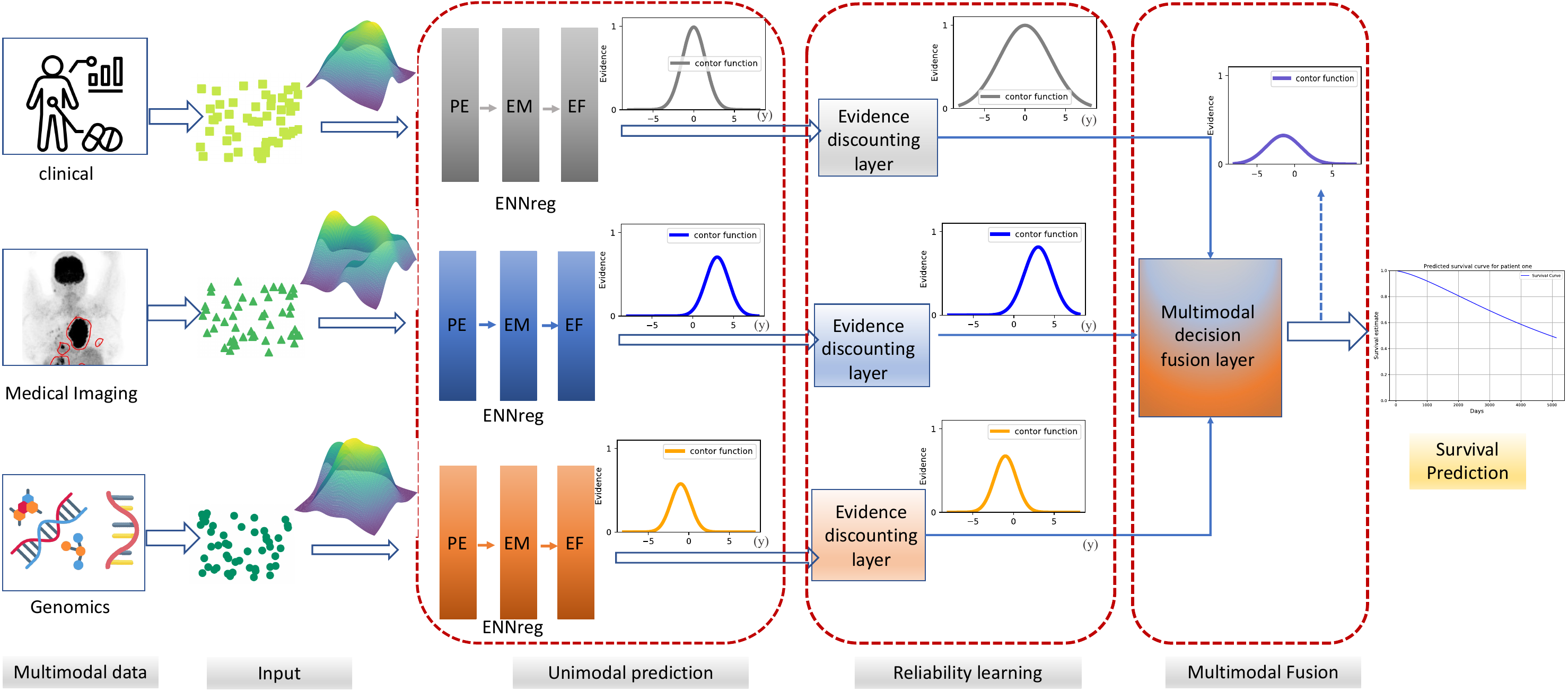}
\caption{Overview of the evidential multimodal survival fusion (EsurvFusion) model. It is composed of several unimodal ENNreg modules that predict survival evidence at the modality level, along with evidence discounting layers that learn the reliability of each modality, and a multimodal decision fusion layer that combines evidence from all modalities. PE is the prototype embedding layer, EM is the evidence mapping layer, and EF is the intermodality prototype-based evidence fusion layer. We visualized the contour functions of the output from each module to better illustrate changes in prediction evidence. }
   \label{fig: framework}
\end{figure*}

\subsection{Multimodal survival analysis}

Multimodal survival analysis is necessary for accurate cancer prognosis with available radiology, histology, omics, and clinical records. The fusion of multimodal survival data can also be grouped into early, middle, and late stages. 
Due to high heterogeneity and noise levels across modalities, the early fusion of survival data suffers from unsatisfying performance. Middle fusion mechanisms are more commonly used to learn shared feature space with technologies such as middle-level feature concatenation \cite{vale2021long}, element-wise sum/multiplication \cite{li2022hfbsurv}, bilinear pooling \cite{chen2020pathomic}. Recently, the popularity of deep learning also contributed to the study of multimodal survival analysis by encoding attention mechanisms \cite{chen2021multimodal, jaume2024modeling, pmlr-v235-song24b} in a deep neural network. However, learning an effectively shared feature space needs enough training data with effective ground truth \cite{vale2021long, chen2021multimodal}. Given that the collection of patients at risk with corresponding survival labels is expensive, and the survival dataset, especially the multimodal survival database, is usually small scale compared to other tasks. Middle fusion-based multimodal survival methods on such small datasets may result in model overfitting and unsatisfying performance. For late fusion, ensemble methods such as Random Survival Forests \cite{ishwaran2008random}, stacking \cite{rahman2023bio}, and gradient boosting \cite{karami2021predicting} are commonly applied to generate multiple decision outputs, which are then aggregated to estimate survival probabilities or cumulative hazard functions. While this approach enhances transparency in decision-making, it requires careful handling of data heterogeneity, censoring, and modality-specific contributions.

%% file: sec/3_method.tex
\newcommand{\reels}{\mathbb{R}}
\newcommand{\tT}{\widetilde{T}}
\newcommand{\tN}{\widetilde{N}}
\newcommand{\tF}{\widetilde{F}}
\newcommand{\tX}{\widetilde{X}}
\newcommand{\tY}{\widetilde{Y}}
\newcommand{\tpsi}{\widetilde{\psi}}
\def\loss{\textsf{loss}}
\newcommand{\calL}{{\cal L}}

\section{Proposed method}
\label{sec: method}
The overview of the proposed EsurvFusion model is displayed in Figure \ref{fig: framework}. It is composed of several unimodal ENNreg modules to encode modality-dependent survival predictions with GRFNs, several evidence discounting layers to learn modality-level reliability, and a multimodal decision fusion layer to combine multimodal decisions. We will introduce the details in the rest of this Section.

\subsection{Preliminaries}
\paragraph{Gaussian random fuzzy number (GRFN)}
A Gaussian fuzzy number (or fuzzy interval) GFN$ \sim N(m, h)$ is a normal fuzzy subset of the real line with membership function $x\mapsto \exp(-\frac{h}{2}(x-m)^2)$, where $m\in \reels$ is the mode and $h\ge 0$ is the precision \cite{denoeux2023reasoning}. In \cite{denoeux2021belief}, random fuzzy sets (RFSs) are used as a model of unreliable and fuzzy evidence. A GRFN \cite{denoeux2023reasoning} $\tY$ can be regarded an RFS defined by a GFN, whose mode $M$ is a Gaussian random variable $M\sim N(\mu, \sigma^2)$ \cite{ibragimov2012gaussian}. A GRFN can therefore be defined by $\tY \sim \tN(\mu, \sigma^2, h)$ with three parameters $\mu$, $\sigma^2$ and $h$. With $\tilde{Y}$, we can calculate different types of evidence on the real line under the DS framework, e.g., the contour function of $\tilde{Y}$ at $y$ is given by 
\begin{equation}
pl_{\tilde{Y}}(y) = \frac{1}{\sqrt{1+h\sigma ^2} }\exp \left(-\frac{h(y-\mu)^2}{2(1+h\sigma^2)} \right). 
\end{equation}
The family of GRFNs is closed under the generalized product-intersection combination rule $\boxplus$ \cite{denoeux2023reasoning}, allowing the combination of multiple uncertain or fuzzy values while remaining within the space of GRFNs. An example usage of $\boxplus$ to combine evidence is provided below. 

\paragraph{Evidential time-to-event prediction}
In \cite{huang2024evidential, huang2024evidentialtimetoevent}, Huang \ et al proposed a three-layer evidential time-to-event prediction model (ENNreg) based on GRFNs. Assuming $K$ data distribution prototypes are available. The prototype embedding (PE) layer computes the distances between the input vector $x$ and prototype $p_k$ with a positive scale parameter $\gamma_k$: $s_k(x)=\exp(-\gamma_k^2 \| x-p_k \|^2)$. The evidence mapping (EM) layer computes a GRFN $\tY(x) \sim 
\tN(\mu_k (x), \sigma_k^2, s_k(x)h_k)$ for prototype $p_k$, $\sigma_k^2$ and $h_k$ are variance and precision, and $\mu_k(x)=\beta_{k}x +\beta_{k0}$, where $\beta_{k}$ is a vector of coefficients and $\beta_{k0}$ is a scalar parameter. 
The evidence fusion (EF) layer combines evidence from the $K$ prototypes using the following generalized product-intersection combination rule $\boxplus$ \cite{denoeux2023reasoning}
 \begin{equation}
 \begin{matrix}
\mu(x)=\frac{\sum_{k=1}^{K} s_k(x)h_k\mu_k(x)}{\sum_{k=1}^{K} s_k(x)h_k}, 
\sigma^2(x)=\frac{\sum_{k=1}^{K}s^2_k(x)h^2_k\sigma^2_k}{(\sum_{k=1}^{K} s_k(x)h_k)^2},\\
h(x)=\sum_{k=1}^{K}s_k(x)h_k.
\end{matrix}
\label{eq: comb}
\end{equation}
Here $\mu(x)$ is the most plausible time-to-event prediction, $\sigma^2(x)$ is the aleatory uncertainty regarding data randomness, and $h(x)$ reflects the epistemic uncertainty of the prediction. ENNreg outputs a GRFN $\tY(x)\sim\tN(\mu(x),\sigma^2(x),h(x))$.

\subsection{Reliability learning}
In most multimodal fusion techniques, each modality is treated as equally reliable. However, due to limitations in data quality and uncertainty about the optimal analysis model, the reliability of information across different modalities varies. In possibility theory, a possibility discounting that takes into account the reliability of evidence was introduced \cite{yager1992considerations}. Recently, Denoeux extended the discounting idea into GRFNs to correct the reliability of prediction \cite{denoeux2024combination}. Inspired by this idea, we correct the reliability of data at the modality level using a possibility discounting approach. Let $\Tilde{Y} \sim \Tilde{N} (\mu, \sigma, h)$, where $M \sim N(\mu, \sigma^2)$ is the mode, the possibility of discounting of $\Tilde{Y}$ can be defined as 
\begin{equation}
   \Tilde{Y}^r= \left [ \exp (-\frac{h}{2}(x-M)^2 \right ]^r =\exp (-\frac{rh}{2} (x-M)^2 ),
\end{equation}
where $0 \le r \le 1$ is the discounting parameter and will be learned based on training data. The discoutned GRFN is expressed as $\Tilde{Y}^r \sim \Tilde{N} (\mu, \sigma, rh)$. 

\subsection{Multimodal decision fusion}
For discounted evidence $\Tilde{Y_i}^{r_{i}}\sim \Tilde{N_i} (\mu_i, \sigma_i, r_i h_i), i=1, \cdots, T$ from $T$ input modalities, we take them as dependent pieces of evidence and use the generalized product-intersection combination rule $\boxplus$ defined in \eqref{eq: comb} to combine them, i.e., 
\begin{equation}
\mu_{f}=\frac{\sum_{i=1}^{T}r_{i}h_{i}\mu_{i}}{\sum_{i=1}^{T}r_{i}h_{i}},
\sigma^2_{f}=\frac{\sum_{i=1}^{T}{r_{i}^{2}h_{i}}^{2} \mu_{i}^{2} }{\sum_{i=1}^{T}{r_{i}}^2{h_{i}}^2},
h_{f}=\sum_{i=1}^{T}r_{i}h_{i}. 
\end{equation}
The multimodal decision fusion can therefore, be expressed as  $\tilde{Y}_{1}^{r_{1}} \boxplus\dots \boxplus\tilde{Y}_{T}^{r_{T}}$ and the finial output is a GFRN $\tF\sim \tilde{N}(\mu_{f},\sigma^2_{f}, h_{f})$, with $\mu_{f}$ the final prediction of the most plausible survival time, $\sigma^2_{f}$ and $h_{f}$ the corresponding aleatory and epistemic uncertainty regarding this prediction.

\subsection{Framework optimization}
Instead of optimizing the model based on the negative partial likelihood, as in conventional survival analysis methods \cite{cox1972regression, katzman2018deepsurv, kvamme2019time}, we optimize the proposed model by minimizing the negative likelihood of two types of evidence, i.e., $Bel_{\tY} $ and $Pl_{\tY}$ that corresponding to the degrees of belief $Bel$ and plausibility $Pl$ given a GRFN that the survival time taking value in a time interval $\left[y_{1}, y_{2}\right]$. 
The detailed formula to calculate $Bel_{\tY} $ and $Pl_{\tY}$ is presented in \cite{huang2024evidentialtimetoevent}. 
Here we explain how to optimize the model with $Bel_{\tY} $ and $Pl_{\tY}$. We consider both unimodal and multimodal optimization and optimize the model with a hybrid multimodal loss function.

\paragraph{Observation transformation} 
The original GRFN takes value in the real line, while the survival time is always positive, we first follow \cite{huang2024evidential} to transform the survival time observation with $t'=\log(t)$ to let GRFN compatible for positive real line. Moreover, to account for the long-tail problem of log transform, we introduced the Yeo-Johnson transformation \cite{weisberg2001yeo} to transform data with approximate normality and further reduce skewness. The final transformed survival time observation is denoted as $y= Y^{J} (\log(t))$, where $J$ is a transformation hyperparameter estimated through maximum likelihood using \emph{Power transform} encoded in the scikit-learn package \cite{pedregosa2011scikit}.

\paragraph{Unimodal optimization} 
Let $\tY^r$ be the output of single modality GRFN, $y$ the transformed time observation, and $D$ a binary censoring indicator. For a given input, if $D=1$, the observation is not censored, meaning the true survival time lies within the range $y\pm \epsilon$, $\epsilon$ is a small positive number. We can therefore optimize the evidence in $ [y-\epsilon, y+\epsilon]$ with
{\small
\begin{equation}
\calL_{Bel}(\tY^r,y^D) = -D\ln Bel_{\tY^r}(y_{\pm \epsilon}),
\end{equation}
}
or 
{\small
\begin{equation}
\calL_{Pl}(\tY^r,y^D) =- D\ln Pl_{\tY^r}(y_{\pm \epsilon}),
\end{equation}
}
If $D=0$, the observation is censored, indicating that the true survival time is at least $y$. We then optimize the evidence in $\left[y, +\infty \right]$ with 
{\small
\begin{equation}
\calL_{Bel}(\tY^r,y^D) = -(1-D)\ln Bel_{\tY^r}([y,\infty)),
\end{equation}
}
or 
{\small
\begin{equation}
\calL_{Pl}(\tY^r,y^D) =- (1-D)\ln Pl_{\tY^r}([y,\infty)),
\end{equation}
}
Following \cite{denoeux2023quantifying, huang2024evidential}, we define a loss function considering both the belief and plausibility of evidence with
\begin{equation}
     Loss_{uni}= \lambda \calL_{Bel}(\tY^r,y^D)+(1-\lambda) \calL_{Pl}(\tY^r,y^D),
 \end{equation}
$\lambda$ is a hyperparameter that controls the cautiousness of the prediction (the smaller, the more cautious).  

\paragraph{Multimodal optimization}
For combined $\tF\sim \tilde{N}(\mu_{f},\sigma^2_{f}, h_{f})$, the calculation of $Bel$, $Pl$, as well as the negative log-likelihood, is the same as unimodal GRFN $\tY^r$. The loss function of multimodal output $Loss_{mul}$ is
\begin{equation} 
     Loss_{mul}= \lambda \calL_{Bel}(\tF,y^D)+(1-\lambda) \calL_{Pl}(\tF,y^D).
 \end{equation}
 The final loss function considering both unimodal and multimodal decisions is defined as 
 \begin{equation}
     Loss_{all}= \sum_{i=1}^{M}{ \eta_i Loss_{uni}(i)} + \varphi Loss_{mul},
 \end{equation}
where $\eta$ is a vector of hyperparameters that balances the loss contributions among multimodal sources, and $\varphi$ is a hyperparameter that mainly focuses on the optimization of the reliability coefficient. The design of $ Loss_{all}$ helps to prevent misleading optimization in highly heterogeneous data.


%% file: sec/4_exp.tex
\begin{table*}[htb]
\centering
\caption{Prediction Accuracy (C-index) comparison of EsurvFusion and compared methods in predicting patients' recurrence-free survival time on HECKTOR2022. The best results are in bold, and the second best are underlined. 
}
\scalebox{1}{
\begin{tabular}{c|c|c|c|c|c}
\hline
\multicolumn{1}{c|}{\multirow{2}{*}{Method}} & \multicolumn{2}{c|}{Single modal}  & \multicolumn{3}{c}{Multimodal} \\ 
&Clinical & Radiomics &Early fusion & Middle fusion &Late fusion \\

\hline
Cox &\underline{0.664}$_{\pm0.016}$	&0.598$_{\pm0.026}$    &0.613$_{\pm0.014}$ & -*-&0.672$_{\pm0.024}$\\
RSF      &0.593$_{\pm0.032}$&\underline{0.602}$_{\pm0.030}$	  &0.614$_{\pm0.019}$  &-*-& 0.638$_{\pm0.019}$ \\
Deepsurv & 0.630$_{\pm0.022}$	 & 0.597$_{\pm0.026}$    & \textbf{0.631}$_{\pm0.045}$ &\textbf{0.663}$_{\pm0.033}$ & \underline{0.686}$_{\pm0.031}$\\
Cox-time & \textbf{0.665}$_{\pm0.012}$	 & \textbf{0.623}$_{\pm0.014}$  & \underline{0.621}$_{\pm0.020}$ & 0.580$_{\pm0.057}$ &0.639$_{\pm0.018}$  \\
Deephit &  0.627$_{\pm0.026}$	& 0.565$_{\pm0.041}$   &0.600$_{\pm0.024}$ & 0.417$_{\pm0.064}$ &0.616$_{\pm0.023}$\\
Multisurv  &-*-	& -*-& -*-& \underline{0.618}$_{\pm0.015}$ &-*-\\  
EsurvFusion (Ours) & -*- &-*- &-*- &-*- &  \textbf{0.703}$_{\pm0.020}$  \\
\hline 
\end{tabular}
}
\label{tab: hecktor}
\end{table*}
\section{Experiments}

\subsection{Experimental Settings}

\paragraph{Dataset} 

To test the effectiveness of EsurvFusion, we choose four cancer survival datasets: HECKTOR 2022\footnote{\url{https://hecktor.grand-challenge.org/Data/}}, and BRCA, BLCA, and COADREAD from the Genomic Data Commons (GDC) database \footnote{\url{https:// portal. gdc. cancer. gov/}}. HECKTOR 2022 is an oropharyngeal Head\&Neck cancer dataset that aims to predict recurrence-free survival (RFS). Patients (n=388, censoring rate=0.81) who underwent radiotherapy and/or chemotherapy treatment planning and have RFS labels are considered. The dataset contains combined PET/CT images of the Head\&Neck region with corresponding clinical tabular data. Breast Invasive Carcinoma (BRCA) (n=1032, censoring rate=0.86), Bladder Urothelial Carcinoma (BLCA) (n=392, censoring rate=0.56), and Colon and Rectum Adenocarcinoma (COADREAD) (n=383, censoring rate=0.76) are three cancer datasets that aim to predict the long-term survival time. The three datasets contain tabular clinical data, gene expression (mRNA), DNA methylation (DNAm), etc. According to \cite{vale2021long}, using only clinical and RNA data achieved the best survival accuracy. Therefore, for the three GDC datasets, we use clinical and RNA data as the multimodal input and report the results. 
For all datasets, we use the original data provided for all modalities except for PET/CT images and use hand-crafted radiomic features \cite{andrearczyk2022head} as the input of PET/CT images. 

\paragraph{Implementation Details}
For each cancer data, the data were split randomly into training, validation, and test sets containing, respectively, 60\%, 20\%, and 20\% of the data. The random split operation was conducted five times with five random seeds to reduce differences in results caused by randomness. Finally, the mean prediction and the standard decision were calculated and reported. For the HECKTOR dataset and three GDC datasets, the number of prototypes $K$ was set to 30 and 40, respectively. The parameters $\epsilon$, $\lambda$, $\eta$, and $\varphi$ were tuned by cross-validation. Specifically, $\epsilon$ was set to $1e^{-4} \sigma_{y}$, $\varphi$ to 0.01, and $\eta$ to 1 for all HECKTOR modalities. For the three GDC datasets, $\eta$ was set to 1 for clinical input and 0.5 for other modalities. The learning rate was 0.05, with a batch size of 512. Adam optimizer was used with an early stopping strategy if the validation loss did not decrease in 20 epochs. 

\paragraph{Evaluation Criteria} We choose concordance index (C-index) \cite{antolini2005time} to measure how well the model correctly predicts the order of events, and use integrated Brier score (IBS) and integrated binomial log-likelihood (IBLL) \cite{kvamme2019time} to evaluate the calibration of the estimates, with lower values indicating more calibrated and reliable predictions.

\subsection{Comparison Baselines}
\paragraph{Survival baseline}
We compare our methods to several established survival models: a classical model, CPH \cite{cox1972regression}; a tree-based ensemble model, Random Survival Forests (RSF) \cite{ishwaran2008random}; a neural network version of CPH, DeepSurv \cite{katzman2018deepsurv}; a time-dependent CPH utilizing neural networks, Cox-Time \cite{kvamme2019time}; a probability mass function-based discrete-time model, DeepHit \cite{lee2018deephit}; and a discrete multimodal survival prediction baseline, MultiSurv \cite{vale2021long}, which extracts multimodal data dependently with a fully connected layer and combines them with attention mechanism. We use the following implementations for the survival baselines: CPH from the Python package lifelines\footnote{\url{https://lifelines.readthedocs.io/en/latest/}}, RSF from scikit-survival\footnote{\url{https://scikit-survival.readthedocs.io/}}, and DeepSurv, Cox-Time, and DeepHit from Pycox\footnote{\url{https://pypi.org/project/pycox/}}. MultiSurv was implemented based on \cite{vale2021long}.

\paragraph{Fusion baseline}
We reproduce early, middle, and late fusion approaches for all survival baselines when compatible. For early fusion, we combine multimodal data by concatenating it into a single input. For middle fusion, we use a fully connected layer to extract features for each modality and then combine them with an attention layer. For late fusion, we analyze each modality with an independent model and combine multimodal predictions through average voting. 

\subsection{Survival Prediction Results and Analysis}

\paragraph{Single modal vs. Multimodal}

Table \ref{tab: hecktor} shows the C-index comparison of EsurvFusion and other methods in predicting patients' recurrence-free survival time (HECKTOR2022) when single-modality and multimodality data are applied. Notably, a simple semi-parametric CPH model trained solely on clinical data and a nonparametric RSF model trained solely on radiomic data serve as strong baselines, achieving C-index of 0.664 and 0.602, respectively, which outperform most deep neural network survival methods. This highlights the challenge and risk of training complex deep models with relatively small datasets.

When multimodal data is applied, we observed that not all fusion methods outperform the single-modality baselines. In most cases, early and middle fusion methods perform worse than using clinical data alone. For example, Cox-Time achieved the highest C-index, 0.665 and 0.623, for single clinical and radiomic inputs, respectively. However, the fusion results for Cox-Time is 0.621, 0.580, and 0.639, respectively, for early, middle, and late fusion. This highlights the need for effective integration of heterogeneous and noisy data modalities.
\begin{figure}[htb]
\centering
\subfloat[\label{fig: clinical}Clinical]{\includegraphics[width=0.2\textwidth,trim=1.6cm 1.0cm 1.1cm 2.0cm,clip]{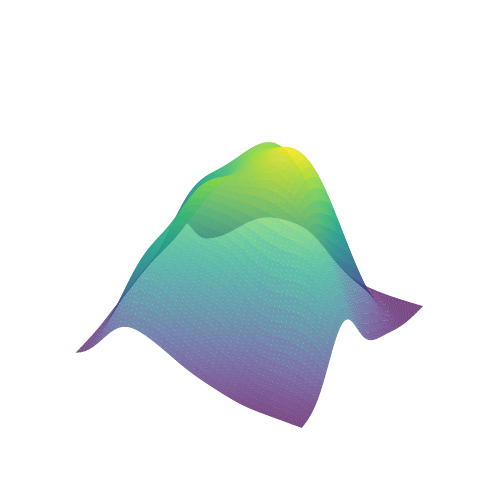}}
\subfloat[\label{fig: radiomic}Radiomic]{\includegraphics[width=0.2\textwidth,trim=1.6cm 1.0cm 1.1cm 3.0cm,clip]{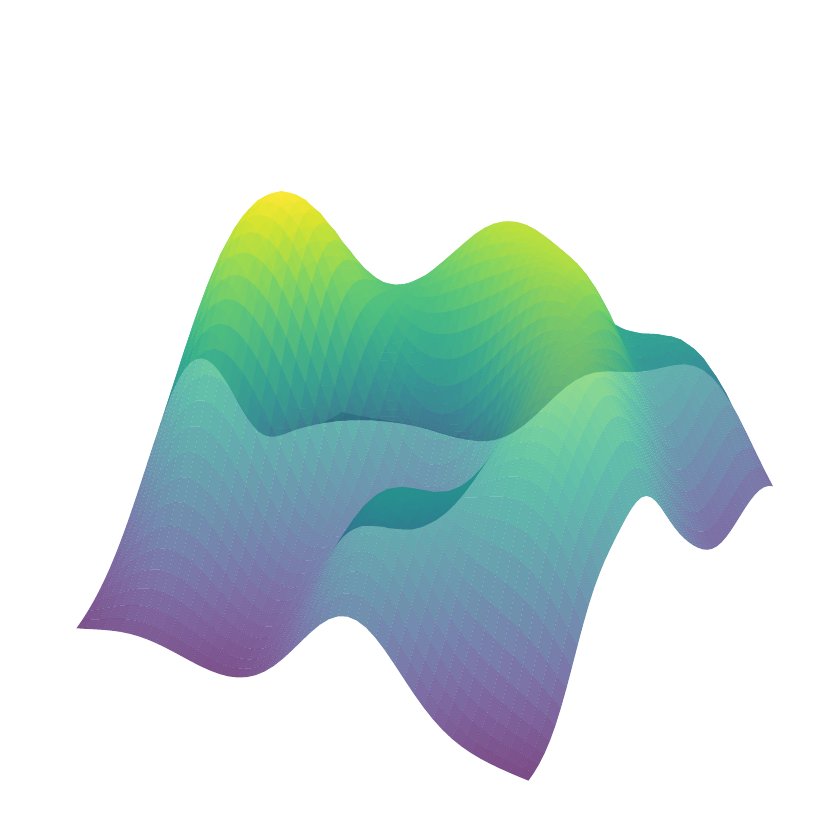}}
\subfloat[\label{fig: concat} Multi-cat]{\includegraphics[width=0.2\textwidth,trim=1.6cm 1.0cm 1.1cm 3.0cm,clip]{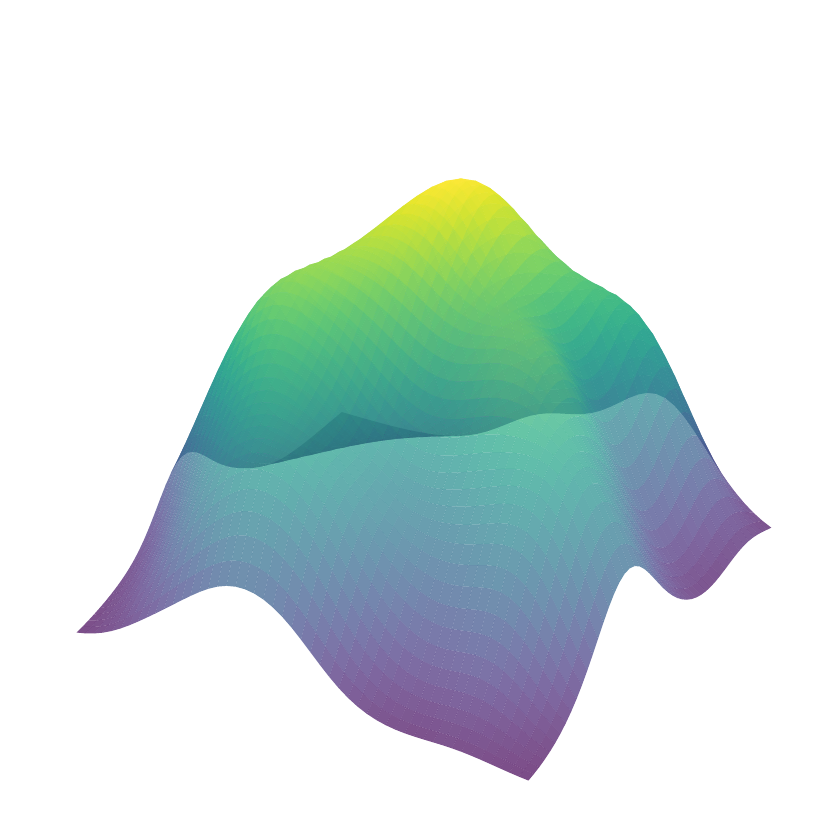}}
\subfloat[\label{fig: attention}Multi-att]{\includegraphics[width=0.23\textwidth,trim=0cm 0.12cm 0cm 0cm,clip]{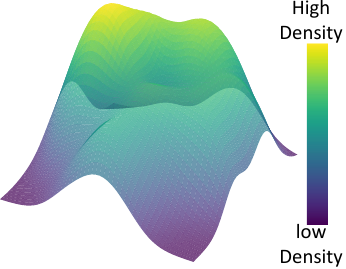}}
\caption{HECKTOR data distribution visualization (via t-SNE): (a) Clinical data distribution. (b) Radiomic data distribution. (c) Combined (clinical and radiomic) data distribution using data concatenation. (d) Combined (clinical and radiomic) feature distribution using an attention mechanism.
The color indicates the distribution density, with yellow representing the highest density. }
\label{fig: feature}
\end{figure}
Figure \ref{fig: feature} shows the data distribution for clinical, radiomic, and combined modalities, with yellow representing the higher distribution density. As shown, both early and middle fusion methods lose some information compared to the single-modality data distributions, and early fusion loses more information than middle fusion. Nevertheless, late fusion stands as the best fusion strategy. Among the late fusion methods, EsurvFusion achieves the best performance with a C-index of 0.703. 

\begin{figure*}[htb]
\centering
\subfloat[\label{fig: cox}CPH (Clinical)]{\includegraphics[width=0.33\textwidth]{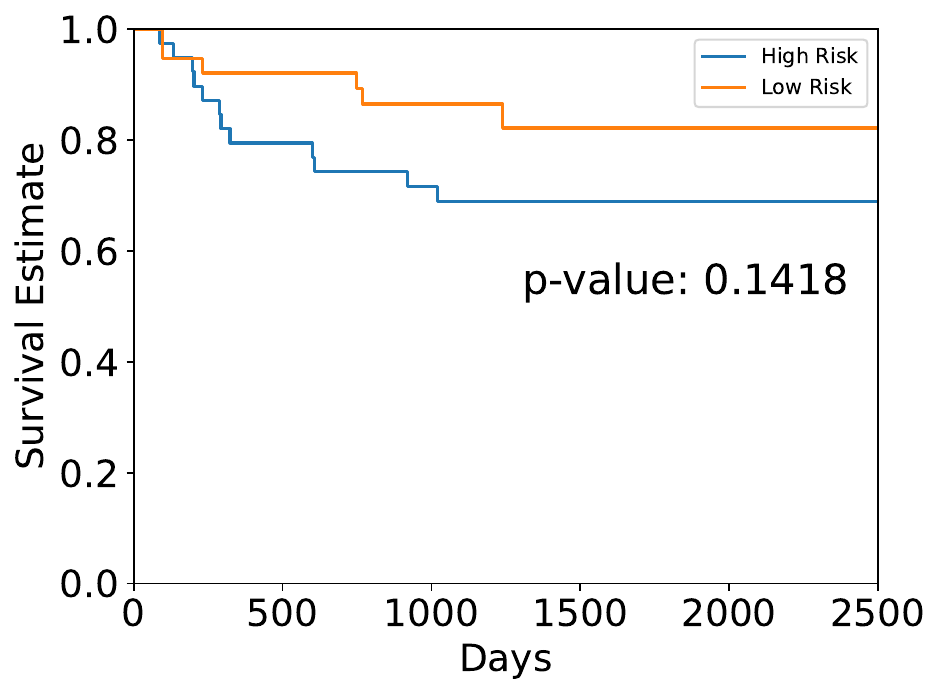}}
\subfloat[\label{fig: rsf} RSF (Radiomic)]
{\includegraphics[width=0.33\textwidth]{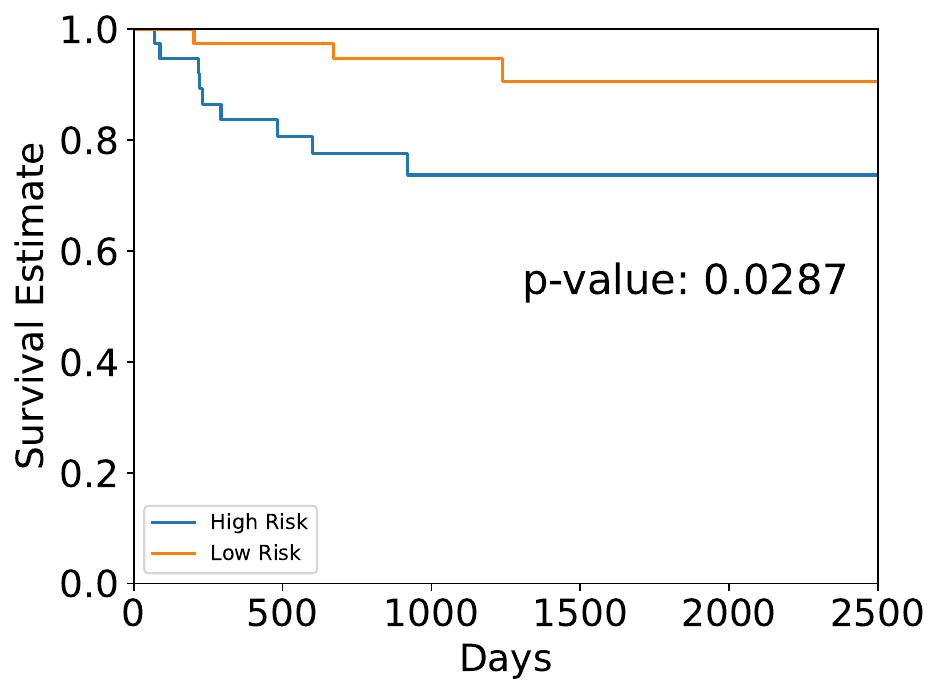}}
\subfloat[\label{fig: deepsurv2}Deepsurv (multi-att)]{\includegraphics[width=0.33\textwidth]{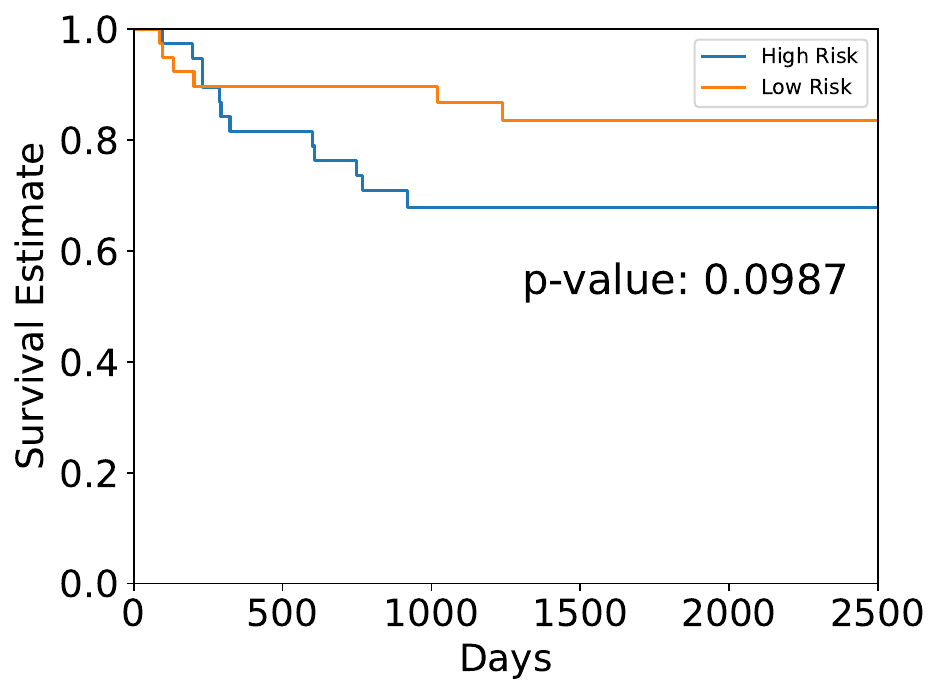}}\\
\subfloat[\label{fig: deepsurv3}Deepsurv (multi-ave)]{\includegraphics[width=0.33\textwidth]{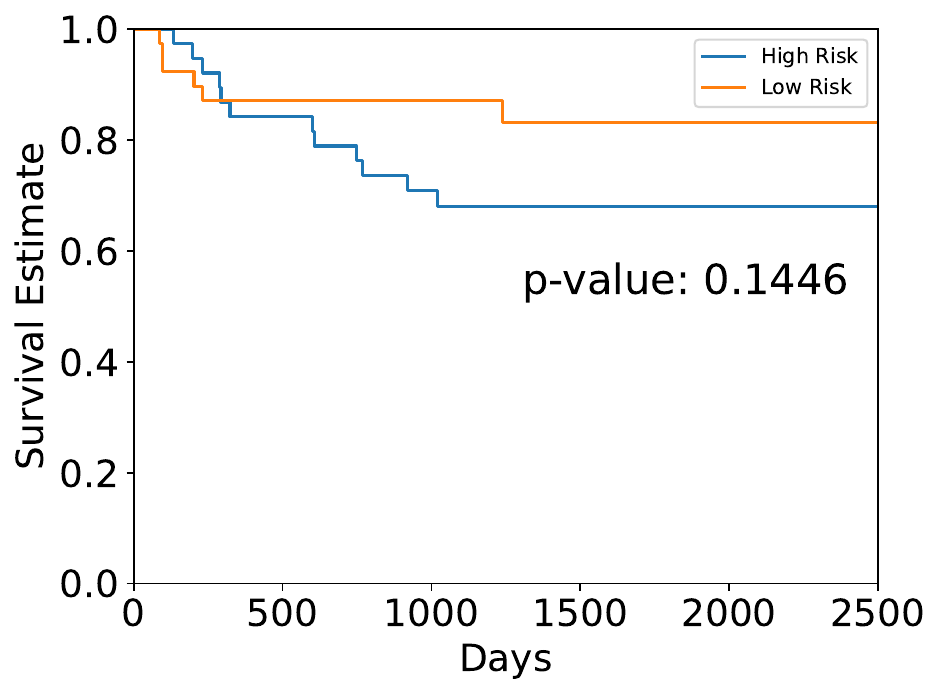}}
\subfloat[\label{fig: survf} \textcolor{red}{EsurvFusion} (multimodal, evidence fusion)]{\includegraphics[width=0.33\textwidth]{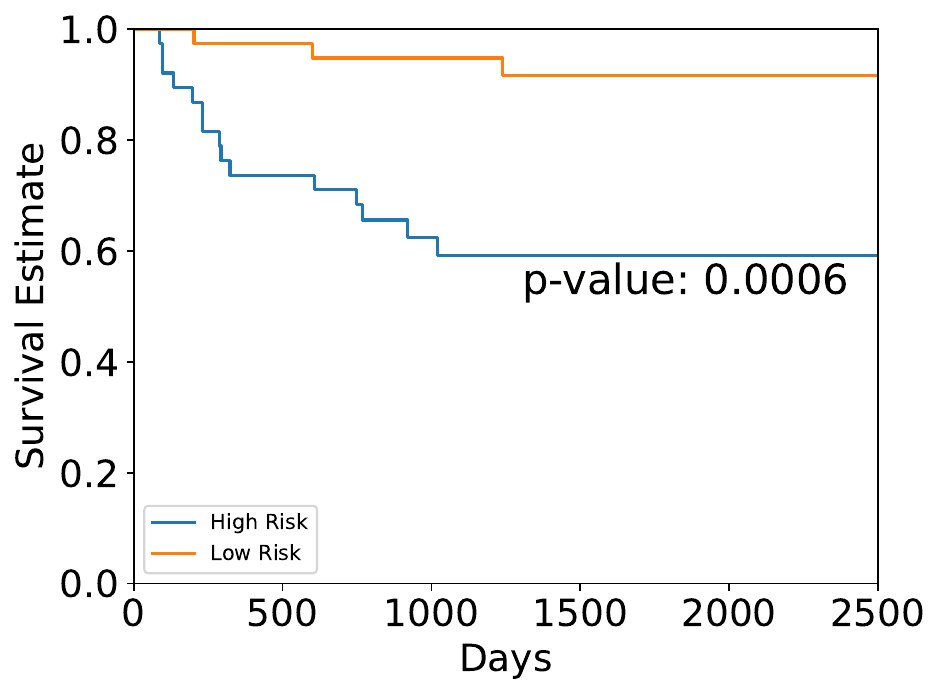}}
\caption{Comparison of the risk stratification performance among different survival methods on HECKTOR2022. High-risk (blue) and low-risk (orange) groups are identified based on the median predicted risk. The Log-rank test was used to determine the statistical significance ($\alpha$ = 0.05).}
\label{fig: KM-estimate}
\end{figure*}
\begin{table*}[htb]
\centering
\caption{Accuracy (C-index) and uncertainty (IBS and IBLL) comparison of EsurvFusion and compared methods in predicting three cancer-specific survival performances: BRCA, BLCA, and COADREAD.  The best results are in bold, and the second best are underlined.
}
\scalebox{0.65}{
\begin{tabular}{@{\hskip -0.2 pt} cl@{\hskip -0.00000001 pt}|c|c|c|c|c|c|c|c|c@{\hskip -0.2 pt}} 
\hline
\multicolumn{2}{c|}{\multirow{2}{*}{Method}} &  \multicolumn{3}{c|}{BRCA} &\multicolumn{3}{c|}{BLCA} 
 & \multicolumn{3}{c}{COADREAD}
 \\
 && C-index ($\uparrow$)& IBS ($\downarrow$) & IBLL ($\downarrow $) &C-index ($\uparrow$)& IBS ($\downarrow$) & IBLL ($\downarrow $) &C-index ($\uparrow$)& IBS ($\downarrow $) & IBLL($\downarrow $) 
 \\
 \hline

\multirow{6}{*}{\rotatebox{90}{Early}} &CPH  &  0.672$_{\pm0.018}$	&0.228$_{\pm0.022}$&1.299$_{\pm0.297}$&0.579$_{\pm0.026}$	&0.344$_{\pm0.023}$	&2.654$_{\pm0.468}$&0.552$_{\pm0.019}$	&0.341$_{\pm0.055}$	&2.688$_{\pm0.626}$\\
&RSF  & 0.718$_{\pm0.004}$	&0.197$_{\pm0.014}$	&0.568$_{\pm0.032}$&0.617$_{\pm0.016}$ &0.225$_{\pm0.009}$ &	0.639$_{\pm0.022}$&0.628$_{\pm0.023}$	&0.216$_{\pm0.007}$	&0.612$_{\pm0.015}$\\
&Deepsurv  &0.598$_{\pm0.027}$	&0.217$_{\pm0.065}$&0.658$_{\pm0.093}$ &0.559$_{\pm0.015}$&0.246$_{\pm0.015}$&0.722$_{\pm0.048}$&0.471$_{\pm0.022}$	&0.216$_{\pm0.020}$	&0.678$_{\pm0.039}$\\
&Cox-time &0.538$_{\pm0.019}$ &0.183$_{\pm0.019}$&0.546$_{\pm0.046}$ &0.592$_{\pm0.005}$	&\textbf{0.214}$_{\pm0.030}$	&0.683$_{\pm0.115}$ &0.490$_{\pm0.039}$&\underline{0.196}$_{\pm0.011}$	&0.637$_{\pm0.045}$\\
&Deephit &0.597$_{\pm0.033}$ &0.183$_{\pm0.006}$&0.545$_{\pm0.018}$ &0.558$_{\pm0.022}$&0.241$_{\pm0.004}$&0.689$_{\pm0.010}$ &0.442$_{\pm0.027}$ &0.212$_{\pm0.011}$	&0.620$_{\pm0.029}$\\
\hline 
\hline 
\multirow{5}{*}{\rotatebox{90}{Middle}}&Deepsurv  & 0.615$_{\pm0.015}$	&0.189$_{\pm0.010}$	& 0.553$_{\pm0.022}$  &0.656$_{\pm0.018}$ &0.226$_{\pm0.011}$	&0.651 $_{\pm0.029}$ &0.556$_{\pm0.042}$	&0.217$_{\pm0.006}$	&0.619$_{\pm0.013}$\\
&Cox-time   &0.660$_{\pm0.030}$	&	0.195	$_{\pm0.009}$	&0.566$_{\pm0.022}$		&0.655$_{\pm0.016}$	&0.224$_{\pm0.005}$	&\underline{0.636}$_{\pm0.015}$		&0.633$_{\pm0.029}$	&\textbf{0.192}$_{\pm0.012}$	&\textbf{0.564}$_{\pm0.028}$\\
&Deephit   & 0.667$_{\pm0.018}$	&0.229$_{\pm0.011}$	&0.649$_{\pm0.028}$		&0.583$_{\pm0.055}$	&0.380$_{\pm0.013}$	&1.095$_{\pm0.033}$	&0.635$_{\pm0.023}$	&0.286$_{\pm0.010}$	&0.824$_{\pm0.027}$\\
&Multisurv    & 0.527$_{\pm0.028}$	&	0.324$_{\pm0.071}$	&1.537$_{\pm0.534}$	&0.527$_{\pm0.028}$	&0.324$_{\pm0.071}$	&1.537 $_{\pm0.534}$		&0.561$_{\pm0.027}$	&0.281$_{\pm0.006}$	&0.970$_{\pm0.061}$\\   
\hline 
\hline 
\multirow{7}{*}{\rotatebox{90}{Late}}& CPH 	&0.723$_{\pm0.019}$	&0.174$_{\pm0.012}$	&0.760$_{\pm0.099}$	&0.625$_{\pm0.009}$&0.315$_{\pm0.009}$	&2.586	$_{\pm0.324}$		&0.625$_{\pm0.009}$	&0.315$_{\pm0.009}$	&2.586$_{\pm0.342}$\\
&RSF 	&0.745$_{\pm0.018}$	&0.170$_{\pm0.007}$	&0.506$_{\pm0.015}$		&0.660$_{\pm0.007}$	&0.220$_{\pm0.005}$	&\textbf{0.626}	$_{\pm0.013}$		&\underline{0.660}$_{\pm0.007}$	&0.220$_{\pm0.005}$	&0.626$_{\pm0.013}$\\
&Deepsurv &0.746$_{\pm0.011}$	&\underline{0.162}$_{\pm0.012}$	&\underline{0.494}$_{\pm0.039}$	&\underline{0.687}$_{\pm0.013}$&0.243$_{\pm0.010}$	&0.729	$_{\pm0.052}$	&0.614$_{\pm0.038}$	&0.227$_{\pm0.009}$ &0.687$_{\pm0.035}$ \\
&Cox-time 	&\underline{0.751}$_{\pm0.013}$	&0.254$_{\pm0.010}$	&0.723$_{\pm0.028}$		&0.656$_{\pm0.023}$	&0.295$_{\pm0.021}$	&1.005$_{\pm0.113}$		&0.656$_{\pm0.023}$	&0.295$_{\pm0.021}$	&1.005$_{\pm0.113}$\\
&Deephit &0.628$_{\pm0.025}$	&0.206$_{\pm0.003}$	&0.608$_{\pm0.013}$	&0.640$_{\pm0.021}$&0.249$_{\pm0.007}$&0.741$_{\pm0.027}$		&0.640$_{\pm0.021}$	&0.249$_{\pm0.007}$	&0.741$_{\pm0.027}$\\  
&EsurvFusion& \textbf{0.782}$_{\pm0.013}$	&\textbf{0.152}$_{\pm0.008}$	&\textbf{0.465}$_{\pm0.020}$ & \textbf{0.690}$_{\pm0.016}$	&\underline{0.218}$_{\pm0.002}$ &0.643$_{\pm0.015}$ &\textbf{0.715}$_{\pm0.024}$	&\underline{0.196}$_{\pm0.006}$ &\underline{0.580}$_{\pm0.016}$\\  
\hline 
\hline 
\end{tabular} 
}

\label{tab: GDC}
\end{table*}
For survival analysis in human cancers, an important clinical evaluation criterion is the ability to stratify patients into high and low-risk subgroups for personalized treatment. Following \cite{jaume2024modeling}, patients with a risk higher than the median of the entire cohort are assigned as high risk (blue), and patients with a risk lower than the median are assigned low risk (orange). A log-rank test based on the Kaplan-Meier survival curve \cite{kleinbaum2012kaplan} was used to assess whether the groups showed significant differences in clinical outcomes. Better prognosis prediction performance comes with a lower p-value in the log-rank test. 
As we can see from Figure \ref{fig: KM-estimate}, EsurvFusion stands out as the best model that can correctly stratify low and high-risk groups with high significant differences (p-value=0.0006).


\paragraph{Early vs. Middle vs. Late fusion}
Table \ref{tab: GDC} presents the results of EsurvFusion and the three fusion methods integrated with compatible baselines and evaluated at three cancer datasets: BRCA, BLCA, and COADREAD, using clinical and RNA data as multimodal inputs. The compared results also show that late fusion strategies consistently outperform early and middle fusion across all baselines. We take BRCA as an example to explain the results.

In early fusion, the best performance was achieved by RSF (0.718 C-index). This observation highlights the high heterogeneity between clinical and RNA data, as RSF is a tree-based ensemble method and is well-suited for handling such variability. Meanwhile, the three neural network models, Deepsurv, Cox-Time, and DeepHit, yielded a C-index of 0.598, 0.538, and 0.597, respectively, performing worse than RSF. When fusing multimodal data using an attention layer, these methods showed improved performance (C-index of 0.615, 0.660, and 0.667) compared to early fusion. We attribute this improvement to enhanced information interaction among multimodal data. 
Nonetheless, their performance still falls short of RSF with early fusion, highlighting the challenge of learning complex models on relatively small datasets with high data heterogeneity.
It is surprising to see that late fusion methods outperformed all early and middle fusion baselines. This finding further claims the necessity of handling data heterogeneity, such as analyzing multimodal data dependently. Among the late fusion methods, EsurvFusion achieved the best performance.

\paragraph{Comparisons with the State-Of-The-Art}
A perfect multimodal survival analysis model should be both accurate and reliable, i.e., achieving a high C-index along with low IBS and IBLL. Models with low IBS and IBLL but a low C-index cannot be considered effective predictors. As shown in Table \ref{tab: GDC}, EsurvFusion yields promising prediction accuracy, outperforming all fusion methods across all the baselines. For the BRCA dataset, the second-best C-index (0.751) was achieved by Cox-Time with late fusion. However, its calibration performance was less satisfactory, with an IBS of 0.254 and an IBLL of 0.723. Compared with Cox-Time, EsurvFusion improves the C-index by 3.1\% and reduces IBS and IBLL by 10.2\% and 25.8\%, respectively. In the BLCA dataset, DeepSurv with late fusion achieved the second-best C-index (0.687), while EsurvFusion improved the C-index by 0.3\% and reduced IBS and IBLL by 2.5\% and 8.6\%, respectively. In the COADREAD dataset, RSF with late fusion achieved the second-best C-index (0.6), while EsurvFusion showed a 5.5\% increase in C-index and reductions of 2.4\% and 4.6\% in IBS and IBLL, respectively. We, therefore, conclude that EsurvFusion is robust to various kinds of cancer survival tasks with accurate and reliable prediction performance.

\paragraph{Prediction and Uncertainty Visualization}

\begin{figure}
\centering
\includegraphics[width=0.7\textwidth]{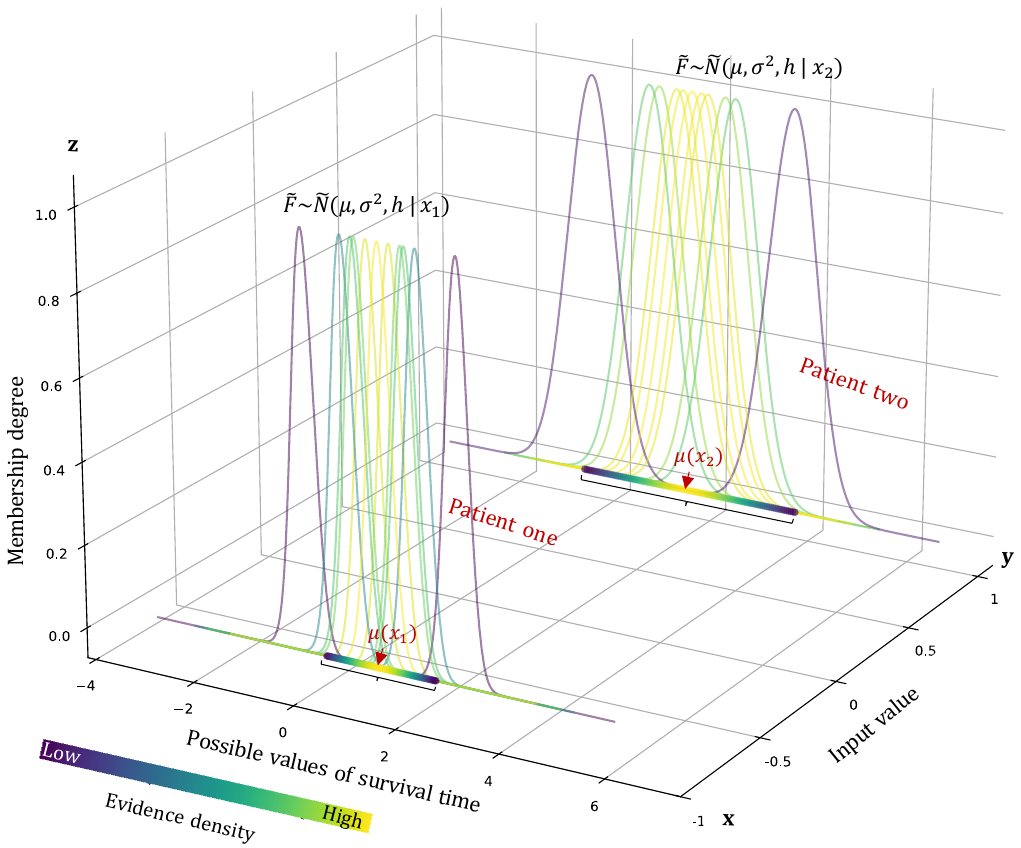}
\caption{The visualization of survival predictions (GRFNs) for two Head\&Neck cancer patients, with input features mapped via t-SNE. For each patient, the transformed possible survival times (GRFNs) are represented by the Gaussian random variable (GRV) $N(\mu, \sigma^2 | x)$ along the x-axis, with the most plausible survival time $\mu(x)$ marked by a red arrow. The less centralized the GRV, the higher the $\sigma^2$, and the higher the aleatory uncertainty. The membership function, tied to precision $h$, is displayed for each GRV. The wider the membership function, the less precision and higher epistemic uncertainty. Color intensity indicates evidence degree.}
\label{fig: prediction}
\end{figure}

Figure \ref{fig: prediction} shows the visualization of survival predictions (GRFNs) for two Head\&Neck cancer patients. The input features $x$ of patients are mapped via t-SNE. For each patient, the transformed possible survival times are represented by the Gaussian random variable $N(\mu, \sigma^2 | x)$, with the most plausible survival time $\mu(x)$ indicated by a red arrow. The membership degree is shown for each possible survival time.
The color intensity indicates the degree of evidence, with yellow representing the highest degree of evidence.
Compared with patient one, patient two shows a larger $\sigma^2$ (represented by less concentrated possible survival times) and lower $h$ (reflected by a wider membership function), suggesting higher aleatoric and epistemic uncertainty in the prediction for patient two. In summary, EsurvFusion allows us to predict survival time directly while providing associated aleatoric and epistemic uncertainty, enhancing prediction interpretability and reliability. 

\subsection{Ablation Study}
We performed an ablation study to assess the effectiveness of reliability learning and multimodal decision fusion. Table \ref{tab: ablation} presents the results of the baseline method ENNreg (B), the baseline with reliability learning (B+R), and the baseline with reliability learning and multimodal fusion (B+R+F), when taking clinical and radiomic as input.
Compared to the baseline, the performance gains from incorporating evidence discounting, e.g., C-index increase from 0.658$\to$0.676 for clinical data and 0.639$\to$0.657 for radiomic data, demonstrating its effectiveness in handling unreliable and noisy data.
Notably, the learned reliability coefficient for clinical data (0.572) is lower than that for radiomic data (0.917), despite the higher C-index achieved by clinical data. This observation reminds the risk of conflating accuracy with reliability, as a high-accuracy model may be overconfident and less robust to data noise, leading to lower reliability. It further supports the effectiveness of assessing information reliability through uncertainty quantification.
With multimodal fusion, we achieved the highest C-index and relatively low IBS and IBLL. The learned reliability coefficients for clinical and radiomic data after fusion increased to 0.856 and 0.966, respectively, indicating that multimodal evidence fusion also facilitates effective information interaction between modalities. These reliability coefficients provide insights into the decision-making process for multimodal inputs, making the process transparent and interpretable. 

\begin{table}
\centering

\caption{Ablation study on reliability learning (R) and multimodal decision fusion (F) with clinical and radiomic data.}

\begin{tabular}{c|c|c|c|c|c}
\hline
Module &   Input  &Reliability &\multicolumn{3}{c}{BRCA} \\
 \cline{4-6}
 &Modality&($r$) & C-index ($\uparrow$)& IBS ($\downarrow$) & IBLL ($\downarrow$) \\
\hline
\multirow{2}{*}{B}  & clinical & -*-&0.658$_{\pm0.015}$ &0.148$_{\pm0.008}$& 0.469$_{\pm0.019}$ \\
\cline{2-6}
 & radiomic &-*-& 0.639$_{\pm0.028}$ & 0.168$_{\pm0.018}$ & 0.520$_{\pm0.047}$ \\  

\hline
\multirow{2}{*}{B+R}    & clinical &0.572$_{\pm0.065}$& 0.676$_{\pm0.016}$ & 0.148$_{\pm0.006}$ &0.468$_{\pm0.014}$\\ 
\cline{2-6}
   & radiomic &0.917$_{\pm0.001}$& 0.657$_{\pm0.026}$ & \textbf{0.138}$_{\pm0.009}$ &\textbf{0.449}$_{\pm0.026}$ \\ 
\hline
B+R+F & clinical &0.856$_{\pm0.017}$& \multirow{2}{*}{\textbf{0.703}$_{\pm0.020}$} & \multirow{2}{*}{\underline{0.141}$_{\pm0.008}$}  &\multirow{2}{*}{\underline{0.452}$_{\pm0.025}$}\\ 
(Final) &\&radiomic &0.966$_{\pm0.003}$ &&&  \\
\hline

\end{tabular}
\label{tab: ablation}
\end{table}

%% file: sec/5_conclu.tex
\section{Conclusion}

In this paper, we proposed a novel multimodal evidential survival fusion model, named EsurvFusion, designed to effectively combine high heterogeneity multimodal survival data at the decision level, incorporating both data and model uncertainty in continuous predictions. Modality-specific reliability is modeled through a possibility discounting operation and corrected before fusion. Final survival predictions are derived through an evidence-based multimodal fusion layer. We evaluated EsurvFusion on four multimodal cancer survival datasets, assessing both accuracy and reliability. The experimental results demonstrated its promising prediction performance along with transparent decision-making and interpretable predictions. For future work, we plan to extend it to large-scale datasets with deep learning, further exploring the advantage of uncertainty quantification in multimodal regression tasks.